\documentclass[pdflatex,sn-mathphys-num,iicol]{sn-jnl}

\usepackage{graphicx}
\usepackage{subcaption}
\usepackage{multirow}
\usepackage{amsmath,amssymb,amsfonts}
\usepackage{amsthm}
\usepackage{mathrsfs}
\usepackage[title]{appendix}
\usepackage{xcolor}
\usepackage{textcomp}
\usepackage{manyfoot}
\usepackage{booktabs}
\usepackage{algorithm}
\usepackage{algorithmicx}
\usepackage{algpseudocode}
\usepackage{listings}
\usepackage{array}
\usepackage{url}

\graphicspath{{img/}}

\theoremstyle{thmstyleone}

\theoremstyle{thmstyletwo}

\theoremstyle{thmstylethree}

\newcommand{\wf}{weighted-F1}
\newcommand{\mf}{macro-F1}
\newcommand{\xgaf}{XGAF}

\begin{document}

\title{SHAP-Weighted Cross-Modal Expert Fusion for Emotion and Sentiment Recognition: Evidence and Limits}

\author*[1]{\fnm{Adis} \sur{Alihodzic}}\email{adis.alihodzic@pmf.unsa.ba}
\author[1]{\fnm{Selma} \sur{Skopljakovic Hubljar}}\email{selma.skoplajkovic@gmail.com}

\affil*[1]{\orgdiv{Department of Mathematical and Computer Sciences}, 
	\orgname{Faculty of Science, University of Sarajevo}, 
	\orgaddress{\city{Sarajevo}, \country{Bosnia and Herzegovina}}}

\abstract{Multimodal emotion and sentiment recognition commonly relies on either early fusion, which concatenates modalities before classification, or late fusion, which combines independently trained unimodal predictors. Early fusion is often accurate but monolithic, whereas late fusion is modular but can lose cross-modal interactions. This paper revisits XAI-guided adaptive fusion (\xgaf), a tree-based mixture of unimodal and cross-modal experts whose sample-level weights are derived from TreeSHAP attribution magnitudes. The study focuses on a methodological issue that becomes important when experts have unequal feature dimensionalities: reducing feature attributions by mean absolute SHAP values can suppress high-dimensional cross-modal experts, while reducing them by summed absolute SHAP values preserves total attribution mass. On MELD 7-class emotion recognition, the proposed sum-abs reduction closes the gap between the cross-modal expert mixture and early fusion across three face-sequence aggregators, with the Transformer variant reaching 0.5983 \wf{} compared with 0.6018 for early fusion and 0.4598 for probability-average late fusion. McNemar testing shows no significant difference between sum-abs \xgaf{} and early fusion on MELD ($p=1.000$), while \xgaf{} remains significantly better than late fusion ($p<0.0001$). On CMU-MOSEI 3-class sentiment recognition, sum-abs \xgaf{} reaches 0.6519 \wf{} compared with 0.6485 for early fusion and 0.5696 for late fusion, with a small but statistically significant improvement over early fusion ($p=0.0452$). An expert-pool ablation indicates that most of the gain comes from adding cross-modal experts, especially the trimodal expert, rather than from rich per-sample routing. A focused three-way ablation further confirms that median-abs SHAP behaves similarly to mean-abs SHAP (0.5682 versus 0.5669 weighted-F1 on MELD), while sum-abs SHAP reaches early-fusion-level performance (0.5957 versus 0.5955). Diagnostic analysis shows that mean-abs and median-abs weights are nearly uniform, whereas sum-abs weights become strongly concentrated on the trimodal expert. The main contribution is therefore not a new state-of-the-art recognition model, but a transparent empirical study of how SHAP attribution reduction, expert dimensionality, and cross-modal expert design affect modular multimodal fusion.}

\keywords{multimodal fusion, emotion recognition, sentiment analysis, explainable artificial intelligence, SHAP, XGBoost, MELD, CMU-MOSEI, mixture of experts}

\maketitle

\section{Introduction}\label{sec:introduction}

Multimodal affective computing attempts to infer human emotion, sentiment, or related affective states from several complementary channels, most commonly text, speech, and visual cues. A sentence may look positive in text but sound sarcastic in speech, or a neutral phrase may become emotionally informative when facial expression is considered. This simple observation explains why fusion is not a minor implementation detail but one of the central design choices in multimodal emotion recognition. Benchmark datasets such as MELD~\cite{poria2019meld} and CMU-MOSEI~\cite{zadeh2018mosei} have made it possible to compare modelling strategies under controlled experimental protocols, but the question of how modalities should be fused remains central. Recent surveys emphasise that modern MER systems must be evaluated not only by accuracy but also by fusion design, robustness to missing or noisy modalities, generalisation across users and datasets, and interpretability of multimodal decisions~\cite{yazici2026survey}. In practice, two simple strategies continue to be widely used. Early fusion concatenates modality-specific feature vectors and trains a single predictor on the joint representation. Late fusion trains separate predictors and combines their output probabilities. Early fusion can exploit cross-modal feature interactions, but it is monolithic and less transparent at the modality level. Late fusion is modular and easy to extend, but it often underuses cross-modal interactions.

A large body of work has proposed more sophisticated fusion mechanisms for multimodal language and affect recognition, including tensor fusion~\cite{zadeh2017tensor}, low-rank multimodal fusion~\cite{liu2018efficient}, cross-modal translation~\cite{pham2019found}, multimodal Transformers~\cite{tsai2019multimodal}, and representation factorisation into modality-invariant and modality-specific components~\cite{hazarika2020misa}. In emotion recognition in conversations, dialogue-aware models further exploit speaker state, discourse structure, graphs, or commonsense knowledge~\cite{majumder2019dialoguernn,ghosal2019dialoguegcn,ghosal2020cosmic,chudasama2022m2fnet}. In the taxonomy of recent MER surveys, these methods mainly occupy hybrid and model-level fusion branches, whereas feature concatenation and probability averaging correspond to early feature-level and decision-level fusion, respectively~\cite{yazici2026survey}. These methods are powerful, but they can be harder to diagnose and deploy when the operational goal is not only maximum benchmark accuracy but also modularity, reproducibility, and interpretable behaviour of modality-specific components.

This paper studies a deliberately simple alternative: a tree-based expert pool composed of unimodal and cross-modal XGBoost classifiers~\cite{chen2016xgboost}. The experts are combined by a sample-dependent gate derived from TreeSHAP attribution magnitudes~\cite{lundberg2017unified}. The motivation is straightforward. If a model's explanation suggests that a given expert has strong attribution mass on a sample, then that expert should receive a higher fusion weight. This idea produces a modular architecture in which text-only, audio-only, face-only, bimodal, and trimodal experts can be inspected separately while still being combined into a single decision.

The central finding of the paper is more nuanced than the original ``adaptive fusion'' motivation. When feature attributions are reduced by the mean of their absolute values, experts with different input dimensionalities are implicitly placed on a per-feature scale. This can normalise away the total attribution mass of high-dimensional cross-modal experts and lead to almost uniform weights. Replacing mean absolute SHAP by summed absolute SHAP restores the scale of cross-modal experts and substantially improves performance. A third natural statistic is the median of the absolute SHAP values. It is more robust to extreme individual attributions and measures a typical feature contribution, but it can also understate experts whose predictive evidence is concentrated in a small number of highly informative features. For this reason, the paper formalises and empirically evaluates mean-, median-, and sum-absolute reductions. The median-abs ablation is important because it is a statistically natural robust alternative, yet the results show that it behaves much closer to mean-abs than to sum-abs. Diagnostic analysis also shows that the improved sum-abs gate does not produce rich per-sample routing on the evaluated datasets. Instead, the gain is largely explained by the availability of cross-modal experts, especially the trimodal expert. This limitation is important: it prevents the method from being overstated, but it also turns the experiment into a useful empirical lesson about attribution-based gating.

The contributions of this paper are as follows:
\begin{enumerate}
    \item A reproducible SHAP-weighted expert-fusion framework is formulated for multimodal emotion and sentiment recognition using unimodal, bimodal, and trimodal XGBoost experts.
    \item The expert-score construction is clarified by formalising and evaluating three natural SHAP reductions: mean-abs, median-abs, and sum-abs. The experiments show that median-abs provides a useful robust baseline, but that preserving total attribution mass with sum-abs is critical when experts have unequal feature dimensionalities.
    \item Experiments on MELD and CMU-MOSEI show that the sum-abs variant matches early fusion on MELD and slightly outperforms it on CMU-MOSEI, while consistently outperforming probability-average late fusion.
    \item Expert-pool ablations identify a compact four-expert variant, denoted \xgaf-Lite, as sufficient to match the full seven-expert pool on MELD.
    \item A diagnostic interpretability analysis is reported, showing that the current SHAP gate mainly produces expert dominance rather than diverse per-sample routing. This negative result is explicitly discussed as a limitation and as guidance for future work.
\end{enumerate}

The paper is intentionally positioned as a fusion-method study rather than as a state-of-the-art dialogue-emotion model. Dialogue context is not modelled, and the goal is to analyse modular expert fusion under a transparent and reproducible classical-machine-learning pipeline.

\section{Related Work}\label{sec:related}

\subsection{Multimodal emotion and sentiment recognition}

MELD extends the EmotionLines corpus with multimodal information from multi-party conversations and is widely used for emotion recognition in conversations~\cite{poria2019meld}. CMU-MOSEI is a large-scale multimodal sentiment and emotion dataset based on opinion videos and aligned language, acoustic, and visual signals~\cite{zadeh2018mosei}. The two datasets differ in task formulation, label structure, and modality quality, making them useful for testing whether a fusion mechanism is dataset-specific.

Fusion methods for multimodal language analysis range from simple concatenation to explicit modelling of high-order interactions. Tensor Fusion Network captures unimodal, bimodal, and trimodal interactions through an outer-product representation~\cite{zadeh2017tensor}. Low-rank multimodal fusion reduces the computational burden of such interactions~\cite{liu2018efficient}. Multimodal Transformer uses cross-modal attention for unaligned multimodal sequences~\cite{tsai2019multimodal}, while MISA separates modality-invariant and modality-specific factors~\cite{hazarika2020misa}. Self-MM further introduces self-supervised multi-task representation learning for multimodal sentiment analysis~\cite{yu2021selfmm}. These models usually target strong end-to-end representation learning, whereas the present paper focuses on a modular tree-based expert pool with explicit diagnostic analysis.

The same survey also highlights three trends that directly motivate the design choices and limitations of this paper: stronger cross-modal fusion, robustness under missing or degraded modalities, and explainability for trustworthy affective computing~\cite{yazici2026survey}. XGAF addresses the explainability and modularity side by using TreeSHAP scores as gate signals, but it does not yet address the missing-modality and noisy-modality scenarios that are increasingly treated as central deployment challenges. This distinction is important because the experiments in this paper use clean pre-extracted features; therefore, the results should be interpreted as a controlled analysis of attribution-based expert weighting rather than as a complete robust MER system.

For conversation emotion recognition, DialogueRNN~\cite{majumder2019dialoguernn}, DialogueGCN~\cite{ghosal2019dialoguegcn}, COSMIC~\cite{ghosal2020cosmic}, and M2FNet~\cite{chudasama2022m2fnet} illustrate the importance of dialogue context and speaker interactions. Because the present work treats each utterance independently after feature extraction, it is not expected to surpass such dialogue-aware methods. This design choice makes the fusion analysis cleaner, but it also defines an important limitation.

\subsection{Mixture of experts and explainable gating}

Mixture-of-experts models combine several specialised predictors through a gating mechanism, and sparse or conditional computation has been used successfully in large-scale neural models~\cite{shazeer2017outrageously}. In multimodal settings, expert specialisation can arise naturally because different modalities have different noise patterns and missingness profiles. However, a gate trained end-to-end can be difficult to interpret, especially when the final model is a deep architecture.

TreeSHAP provides consistent feature attributions for tree ensemble models and is widely used to explain individual predictions~\cite{lundberg2017unified}. This paper uses TreeSHAP not only for post-hoc explanation but also as a gate signal. The idea is simple: an expert receives a higher weight when its absolute attribution mass is larger on the current sample. The experiments show, however, that a post-hoc attribution signal is not automatically a reliable routing signal. The way attributions are reduced to one scalar per expert has a strong influence on the resulting gate.

\section{Method}\label{sec:method}

\subsection{Expert pool}

Let each sample be represented by three modality-specific feature vectors: text $x_t$, voice $x_v$, and face $x_f$. The full modality set is $M=\{t,v,f\}$. For each non-empty subset $S\subseteq M$, an expert $h_S$ may be trained on the concatenated feature vector $x_S$. In the full configuration, the expert pool is
\begin{equation}
\mathcal{E}=\{h_t,h_v,h_f,h_{tv},h_{tf},h_{vf},h_{tvf}\}.
\end{equation}
The paper also evaluates smaller pools, including a unimodal-only pool and a compact pool containing the three unimodal experts plus the trimodal expert. All experts are implemented as XGBoost classifiers. Each expert returns a class-probability vector $p_e(y\mid x)$.

Early fusion is the special case in which only the trimodal classifier $h_{tvf}$ is used. Late fusion is implemented as the unweighted average of unimodal output probabilities. The proposed framework generalises both by allowing a pool of unimodal and cross-modal experts to be combined through SHAP-derived weights. From the perspective of recent MER taxonomies, this design is best viewed as an interpretable expert-level fusion layer rather than an end-to-end model-level fusion architecture: it keeps the representations and experts modular, but it uses sample-dependent attribution evidence to combine their predictions~\cite{yazici2026survey}.

\subsection{SHAP-derived expert scores}

For a sample $x$ and expert $e$, TreeSHAP produces feature-attribution values. For multi-class classification, the implementation aggregates absolute attribution magnitudes over classes and features. Let $d_e$ denote the number of input features of expert $e$ and let $\phi_{e,j}(x)$ denote the attribution contribution associated with feature $j$ after the class-level aggregation. Three natural scalar expert scores can be defined:
\begin{align}
    a^{\mathrm{mean}}_e(x)&=\frac{1}{d_e}\sum_{j=1}^{d_e}|\phi_{e,j}(x)|,
    \label{eq:mean_abs}\\
    a^{\mathrm{median}}_e(x)&=\operatorname{median}_{j=1,\ldots,d_e}|\phi_{e,j}(x)|,
    \label{eq:median_abs}\\
    a^{\mathrm{sum}}_e(x)&=\sum_{j=1}^{d_e}|\phi_{e,j}(x)|.
    \label{eq:sum_abs}
\end{align}
The mean-abs score measures the average attribution per feature. The median-abs score measures the typical attribution and is robust to unusually large individual feature contributions. The sum-abs score measures the total attribution mass carried by the whole expert. This distinction matters because expert-level gating is not the same as ranking individual features. A median score can be useful when a few outlier attributions should not dominate the gate, but it may also undervalue an expert whose evidence is sparse and concentrated in only a few highly informative dimensions. A sum score has the opposite bias: it preserves total evidence, but it can favour high-dimensional cross-modal experts.

When experts have strongly different dimensions, the reduction choice is therefore not merely cosmetic. In the MELD setup, text and voice experts have 768 features, the face expert has 512 features, bimodal experts have 1280 or 1536 features, and the trimodal expert has 2048 features. Mean-abs and median-abs reductions place experts on a typical-per-feature scale, whereas sum-abs reduction compares total attribution mass. The empirical tables in this preprint report mean-, median-, and sum-abs reductions. This makes the reduction analysis more informative: mean-abs and median-abs both operate on a typical-feature scale, whereas sum-abs operates on the total-evidence scale of the whole expert.

\subsection{Temperature-scaled fusion}

Given expert scores $a_e(x)$, the gate assigns a softmax weight
\begin{equation}
    w_e(x;\tau)=\frac{\exp(a_e(x)/\tau)}{\sum_{e'\in\mathcal{E}}\exp(a_{e'}(x)/\tau)},
\end{equation}
where $\tau>0$ is selected on the validation set. The final probability vector is
\begin{equation}
    p(y\mid x)=\sum_{e\in\mathcal{E}} w_e(x;\tau)p_e(y\mid x).
\end{equation}
A low temperature produces concentrated weights, while a high temperature approaches uniform averaging. The temperature is tuned by maximising validation \wf{}.

\begin{algorithm}[t]
\caption{SHAP-weighted cross-modal expert fusion}
\label{alg:xgaf}
\begin{algorithmic}[1]
\Require Training, validation, and test splits with modality features $x_t,x_v,x_f$; expert pool $\mathcal{E}$; temperature grid $\mathcal{T}$; SHAP reduction mode $r\in\{\mathrm{mean},\mathrm{median},\mathrm{sum}\}$.
\For{each expert $e\in\mathcal{E}$}
    \State Train XGBoost classifier $h_e$ on its modality subset.
    \State Compute validation probabilities and TreeSHAP attributions.
\EndFor
\For{each $\tau\in\mathcal{T}$}
    \State Convert SHAP attributions to expert scores using reduction $r$.
    \State Compute softmax expert weights and fused validation probabilities.
    \State Record validation \wf{}.
\EndFor
\State Select $\tau^*$ with the best validation \wf{}.
\State Apply the same experts, reduction mode, and $\tau^*$ to the test set.
\State \Return fused test predictions and diagnostic expert weights.
\end{algorithmic}
\end{algorithm}

\section{Experimental Setup}\label{sec:setup}

\subsection{Datasets and feature representations}

Table~\ref{tab:datasets} summarises the two experimental settings. MELD is used for seven-class emotion recognition, and CMU-MOSEI is used for three-class sentiment recognition. The filtering counts refer to samples for which all required modality features were available.

\begin{table*}[t]
\caption{Datasets and feature dimensions used in the experiments.}
\label{tab:datasets}
\centering
{\footnotesize
\setlength{\tabcolsep}{4pt}
\begin{tabular}{lcccccc}
\toprule
Dataset & Task & Train & Val. & Test & Text & Voice / Face \\
\midrule
MELD & 7-class emotion & 9660 & 1067 & 2525 & 768 & 768 / $15\times512$ \\
CMU-MOSEI & 3-class sentiment & 16326 & 1871 & 4659 & 768 & 74 / 35 \\
\bottomrule
\end{tabular}\par}
\end{table*}

For MELD, text features are BERT-base [CLS] embeddings~\cite{devlin2019bert}, voice features are wav2vec 2.0-based embeddings~\cite{baevski2020wav2vec}, and face features are obtained from a face-emotion pipeline based on frame sampling, face detection, and an Emo-AffectNet-style visual backbone~\cite{ryumina2022emoaff}. Fifteen face-frame embeddings are aggregated by mean pooling, a BiLSTM aggregator, or a Transformer aggregator. These three variants test whether the fusion conclusions depend on the temporal face aggregator.

For CMU-MOSEI, the experiments use preprocessed aligned features with 768-dimensional text vectors, 74-dimensional acoustic vectors, and 35-dimensional visual vectors. The original sentiment labels are mapped to a three-class setting. The purpose of this dataset is not to claim a new state-of-the-art result, but to test whether the SHAP-reduction behaviour observed on MELD transfers to a second multimodal benchmark.

\subsection{Baselines and evaluated variants}

The following methods are compared:
\begin{itemize}
    \item \textbf{Text-only XGBoost}: a strong unimodal baseline using only text features.
    \item \textbf{Early fusion}: one XGBoost classifier trained on the concatenation of all available modality features.
    \item \textbf{Late fusion}: probability averaging of unimodal XGBoost classifiers.
    \item \textbf{XGAFv1}: a unimodal-only Random Forest variant retained as a historical baseline.
    \item \textbf{XGAFv2 mean-abs}: a seven-expert XGBoost pool gated by mean absolute SHAP scores.
    \item \textbf{XGAFv2 sum-abs}: a seven-expert XGBoost pool gated by summed absolute SHAP scores.
    \item \textbf{XGAFv2 median-abs}: a robust variant gated by median absolute SHAP scores, included to test whether a typical-feature statistic improves over mean-abs without allowing high-dimensional experts to dominate as strongly as sum-abs.
    \item \textbf{XGAF-Lite}: a compact four-expert pool with three unimodal experts and one trimodal expert.
\end{itemize}

Performance is reported using accuracy, \wf{}, \mf{}, and Cohen's $\kappa$. Because MELD is class-imbalanced, \wf{} is treated as the main metric. Pairwise prediction differences are evaluated with the exact-binomial version of McNemar's test using the same test instances.

\section{Results}\label{sec:results}

\subsection{MELD results}

Table~\ref{tab:meld_transformer} reports the main MELD results for the Transformer face aggregator. Text-only XGBoost is already a strong baseline, reaching 0.5935 \wf{}. Early fusion improves this to 0.6018. Probability-average late fusion is much weaker, reaching only 0.4598 \wf{}, which confirms that naive late fusion loses useful cross-modal structure. The mean-abs version of XGAFv2 reaches 0.5714 \wf{}, while the sum-abs version improves to 0.5983. The compact XGAF-Lite variant reaches 0.6013 \wf{}, essentially matching early fusion.

\begin{table*}[t]
\caption{MELD test results with the Transformer face aggregator.}
\label{tab:meld_transformer}
\centering
{\footnotesize
\setlength{\tabcolsep}{4pt}
\begin{tabular}{lcccc}
\toprule
Method & Accuracy & \wf{} & \mf{} & $\kappa$ \\
\midrule
Text-only XGB & 0.6257 & 0.5935 & 0.4091 & 0.4292 \\
Early fusion XGB & 0.6313 & \textbf{0.6018} & \textbf{0.4090} & \textbf{0.4392} \\
Late fusion avg. & 0.5541 & 0.4598 & 0.2511 & 0.2083 \\
XGAFv1 unimodal RF & 0.5743 & 0.4862 & 0.2593 & 0.2535 \\
XGAFv2 mean-abs, all cross & 0.6222 & 0.5714 & 0.3747 & 0.3910 \\
XGAFv2 sum-abs, all cross & \textbf{0.6317} & 0.5983 & 0.4051 & 0.4349 \\
XGAF-Lite sum-abs & 0.6309 & 0.6013 & 0.4087 & 0.4384 \\
\bottomrule
\end{tabular}\par}
\end{table*}

Table~\ref{tab:aggregators} shows that the main conclusion is stable across face-sequence aggregators. The sum-abs variant consistently reduces the gap to early fusion and improves over mean-abs by 2.52 to 2.90 percentage points in \wf{}. The three face aggregators are close to each other, suggesting that the fusion behaviour is not driven by a specific face-sequence model. Table~\ref{tab:median_ablation} then adds the explicit median-abs ablation for the Transformer setup.

\begin{table*}[t]
\caption{MELD comparison of mean-abs and sum-abs SHAP reductions across face aggregators.}
\label{tab:aggregators}
\centering
{\footnotesize
\setlength{\tabcolsep}{4pt}
\begin{tabular}{lcccc}
\toprule
Aggregator & Early fusion & XGAF mean-abs & XGAF sum-abs & Gain over mean \\
\midrule
Mean pooling & 0.6029 & 0.5686 & 0.5976 & +0.0290 \\
BiLSTM & 0.6007 & 0.5749 & 0.6001 & +0.0252 \\
Transformer & 0.6018 & 0.5714 & 0.5983 & +0.0269 \\
\bottomrule
\end{tabular}\par}
\end{table*}

\subsection{Mean-, median-, and sum-abs SHAP ablation}

Table~\ref{tab:median_ablation} reports the focused SHAP-reduction ablation for the Transformer face aggregator using the same seven-expert pool. This experiment was added to test whether the median of absolute SHAP values is a useful robust alternative to the mean and the sum. The result is clear: median-abs improves only marginally over mean-abs (+0.13 percentage points in \wf{}), while sum-abs improves over median-abs by 2.74 percentage points and matches the early-fusion reference. The entropy diagnostics also separate the reductions. Mean-abs and median-abs produce almost maximum-entropy weights across the seven experts, indicating nearly uniform weighting. Sum-abs produces much lower entropy and assigns a dominant expert on every test sample. Thus, median-abs is a sensible statistical baseline, but it does not solve the expert-scale problem in this architecture.

\begin{table*}[t]
\caption{Focused SHAP-reduction ablation on MELD with the Transformer face aggregator. The table compares mean-, median-, and sum-absolute reductions using the same XGAFv2 expert pool.}
\label{tab:median_ablation}
\centering
{\footnotesize
\setlength{\tabcolsep}{4pt}
\begin{tabular}{lcccccccc}
\toprule
Method & Accuracy & \wf{} & \mf{} & $\kappa$ & Val. \wf{} & $\tau^*$ & Entropy & Dom. rate \\
\midrule
Early fusion (reference) & 0.6265 & 0.5955 & 0.4022 & 0.4299 & -- & -- & -- & -- \\
XGAFv2 mean-abs & 0.6190 & 0.5669 & 0.3696 & 0.3851 & 0.5276 & 0.7 & 1.9454 & 0.0 \\
XGAFv2 median-abs & 0.6198 & 0.5682 & 0.3713 & 0.3866 & 0.5266 & 2.5 & 1.9459 & 0.0 \\
XGAFv2 sum-abs & \textbf{0.6269} & \textbf{0.5957} & \textbf{0.4022} & \textbf{0.4304} & \textbf{0.5611} & 1.5 & 0.3912 & 1.0 \\
\bottomrule
\end{tabular}\par}
\end{table*}

\subsection{Statistical testing on MELD}

Table~\ref{tab:mcnemar} reports pairwise McNemar tests on the MELD test set for the Transformer aggregator. Early fusion and sum-abs XGAFv2 are statistically indistinguishable ($p=1.000$). Both early fusion and sum-abs XGAFv2 are significantly better than probability-average late fusion ($p<0.0001$). The small difference between text-only and the two strongest multimodal methods is not statistically significant under this test, which again indicates that MELD is strongly text-dominated under the present per-utterance setup.

\begin{table*}[t]
\caption{Pairwise McNemar tests on MELD, Transformer aggregator. Here $b$ denotes samples on which method A is wrong and method B is correct, while $c$ denotes samples on which A is correct and B is wrong.}
\label{tab:mcnemar}
\centering
{\footnotesize
\setlength{\tabcolsep}{4pt}
\begin{tabular}{lrrrr}
\toprule
Comparison & $\Delta$ acc. & $b$ & $c$ & $p_{\mathrm{exact}}$ \\
\midrule
Early vs. XGAF sum-abs & +0.0004 & 36 & 35 & 1.0000 \\
Early vs. late fusion & -0.0772 & 146 & 341 & $<0.0001$ \\
XGAF sum-abs vs. late fusion & -0.0776 & 125 & 321 & $<0.0001$ \\
Early vs. text-only & -0.0055 & 131 & 145 & 0.4340 \\
XGAF sum-abs vs. text-only & -0.0059 & 110 & 125 & 0.3611 \\
\bottomrule
\end{tabular}\par}
\end{table*}

\subsection{CMU-MOSEI cross-dataset validation}

Table~\ref{tab:mosei} reports the corresponding CMU-MOSEI results. Sum-abs XGAFv2 reaches 0.6519 \wf{}, compared with 0.6485 for early fusion and 0.5696 for late fusion. The advantage over early fusion is small, but McNemar testing gives $p=0.0452$. Since the effect size is only 0.34 percentage points in \wf{}, it should be interpreted as evidence that the method is competitive with early fusion rather than as a large practical improvement.

\begin{table*}[t]
\caption{CMU-MOSEI 3-class sentiment results.}
\label{tab:mosei}
\centering
{\footnotesize
\setlength{\tabcolsep}{4pt}
\begin{tabular}{lcccc}
\toprule
Method & Accuracy & \wf{} & \mf{} & $\kappa$ \\
\midrule
Text-only XGB & 0.6649 & 0.6396 & 0.5871 & 0.4369 \\
Early fusion & 0.6729 & 0.6485 & 0.5984 & 0.4497 \\
Late fusion avg. & 0.6280 & 0.5696 & 0.4995 & 0.3350 \\
XGAFv2 mean-abs & 0.6622 & 0.6203 & 0.5604 & 0.4124 \\
XGAFv2 sum-abs & \textbf{0.6798} & \textbf{0.6519} & \textbf{0.5996} & \textbf{0.4592} \\
\bottomrule
\end{tabular}\par}
\end{table*}

\subsection{Expert-pool ablation}

\begin{figure}[t]
\centering
\includegraphics[width=\linewidth]{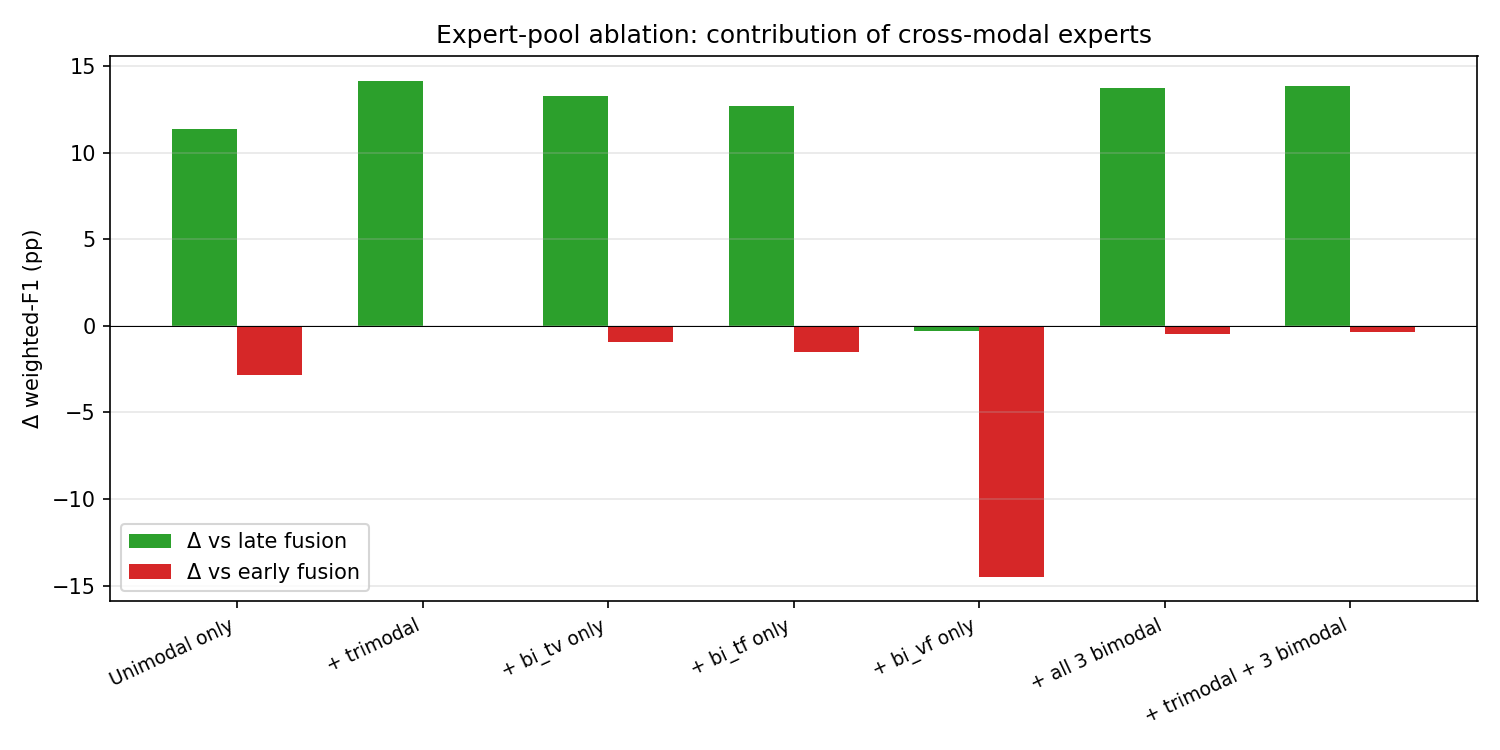}
\caption{Expert-pool ablation on MELD. Positive values relative to late fusion show that cross-modal experts recover much of the performance lost by naive late fusion. The four-expert XGAF-Lite variant is sufficient to match early fusion.}
\label{fig:ablation}
\end{figure}

Table~\ref{tab:ablation} and Fig.~\ref{fig:ablation} show the MELD expert-pool ablation. The unimodal-only pool reaches 0.5736 \wf{}. Adding only the trimodal expert increases performance to 0.6013 \wf{}, which is almost identical to early fusion. Adding all three bimodal experts without the trimodal expert reaches 0.5972, and the full seven-expert pool reaches 0.5983. These results indicate that the main improvement comes from adding cross-modal experts, especially the trimodal expert, rather than from increasing the number of experts.

\begin{table*}[t]
\caption{Expert-pool ablation on MELD, Transformer aggregator, sum-abs reduction.}
\label{tab:ablation}
\centering
{\footnotesize
\setlength{\tabcolsep}{4pt}
\begin{tabular}{lrrrr}
\toprule
Variant & Experts & \wf{} & $\Delta$ early & $\Delta$ late \\
\midrule
Unimodal only & 3 & 0.5736 & -0.0282 & +0.1138 \\
+ trimodal (XGAF-Lite) & 4 & \textbf{0.6013} & -0.0005 & +0.1415 \\
+ text-voice only & 4 & 0.5924 & -0.0095 & +0.1326 \\
+ text-face only & 4 & 0.5870 & -0.0149 & +0.1272 \\
+ voice-face only & 4 & 0.4571 & -0.1448 & -0.0027 \\
+ all three bimodal & 6 & 0.5972 & -0.0046 & +0.1374 \\
+ trimodal + all bimodal & 7 & 0.5983 & -0.0036 & +0.1385 \\
\bottomrule
\end{tabular}\par}
\end{table*}

\section{Interpretability and Diagnostic Analysis}\label{sec:diagnostics}

\subsection{Why mean-abs and sum-abs behave differently}

\begin{figure}[t]
\centering
\includegraphics[width=\linewidth]{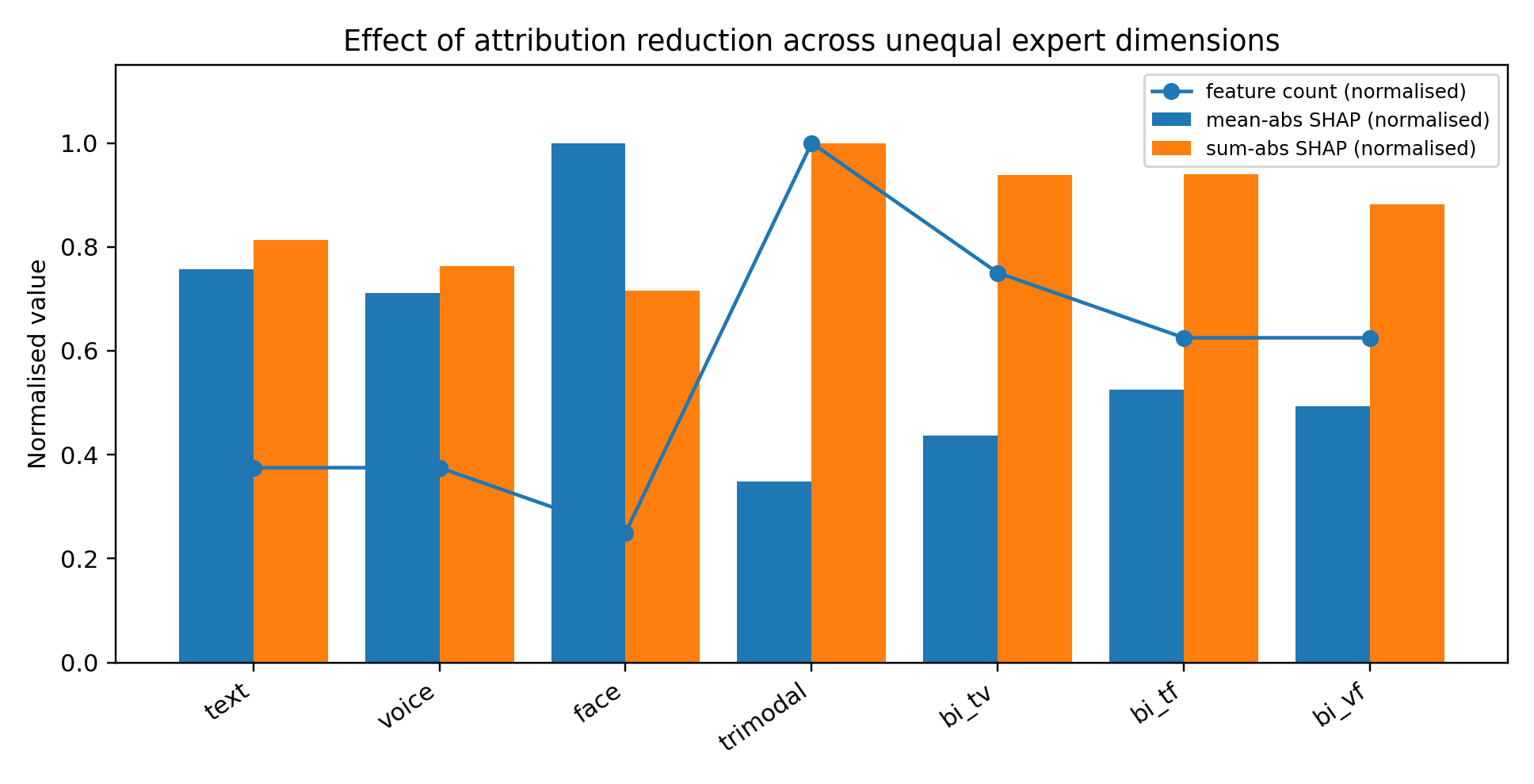}
\caption{Normalised diagnostic comparison of feature counts, mean-abs SHAP scores, and sum-abs SHAP scores on MELD. Mean-abs reduction can make high-dimensional cross-modal experts appear weak on an average-per-feature scale, while sum-abs reduction preserves total attribution mass.}
\label{fig:reduction_diag}
\end{figure}

Figure~\ref{fig:reduction_diag} summarises the source of the reduction difference. The experts have unequal feature dimensionalities, but mean-abs reduction places each expert on an average-per-feature scale. Median-abs goes further toward a robust typical-feature interpretation by ignoring extreme attribution values more strongly than the mean. This is attractive statistically, but the explicit ablation in Table~\ref{tab:median_ablation} shows that it is too conservative for expert gating in the present architecture: it remains close to mean-abs and far below sum-abs in \wf{}. In the MELD run, the trimodal expert has the largest feature count but a relatively low mean attribution value. The sum-abs reduction preserves total attribution mass and therefore gives larger scores to cross-modal experts when their aggregate contribution is large.

\subsection{Expert weights do not show rich adaptive routing}

\begin{figure}[t]
\centering
\includegraphics[width=\linewidth]{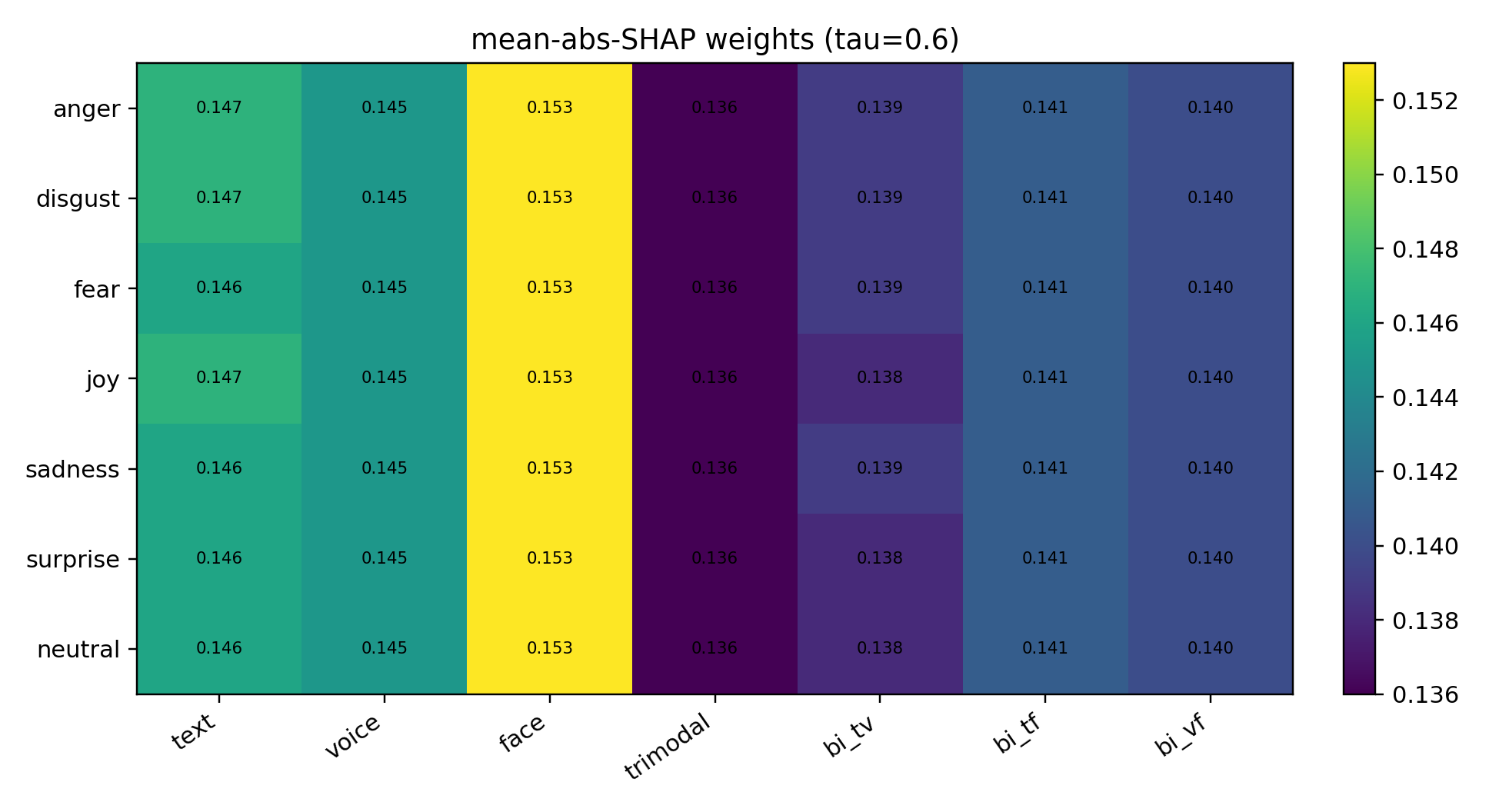}
\caption{Average expert weights per MELD emotion class using mean-abs SHAP reduction. The weights are close to uniform, so the gate behaves similarly to simple averaging.}
\label{fig:weights_mean}
\end{figure}

\begin{figure}[t]
\centering
\includegraphics[width=\linewidth]{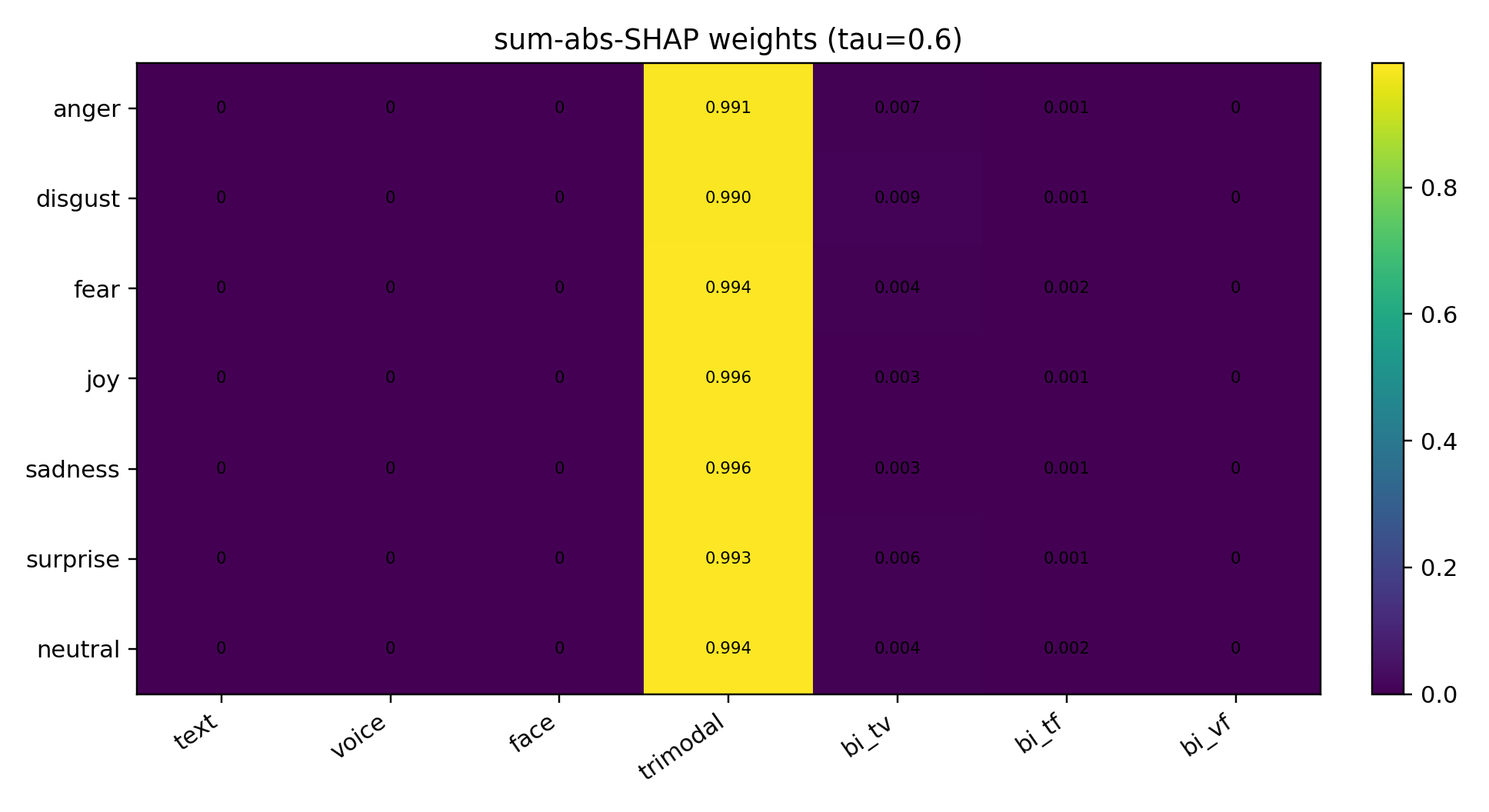}
\caption{Average expert weights per MELD emotion class using sum-abs SHAP reduction. The gate becomes concentrated on the trimodal expert, which explains the recovery of early-fusion-like performance but also shows limited routing diversity.}
\label{fig:weights_sum}
\end{figure}

\begin{figure}[t]
\centering
\includegraphics[width=\linewidth]{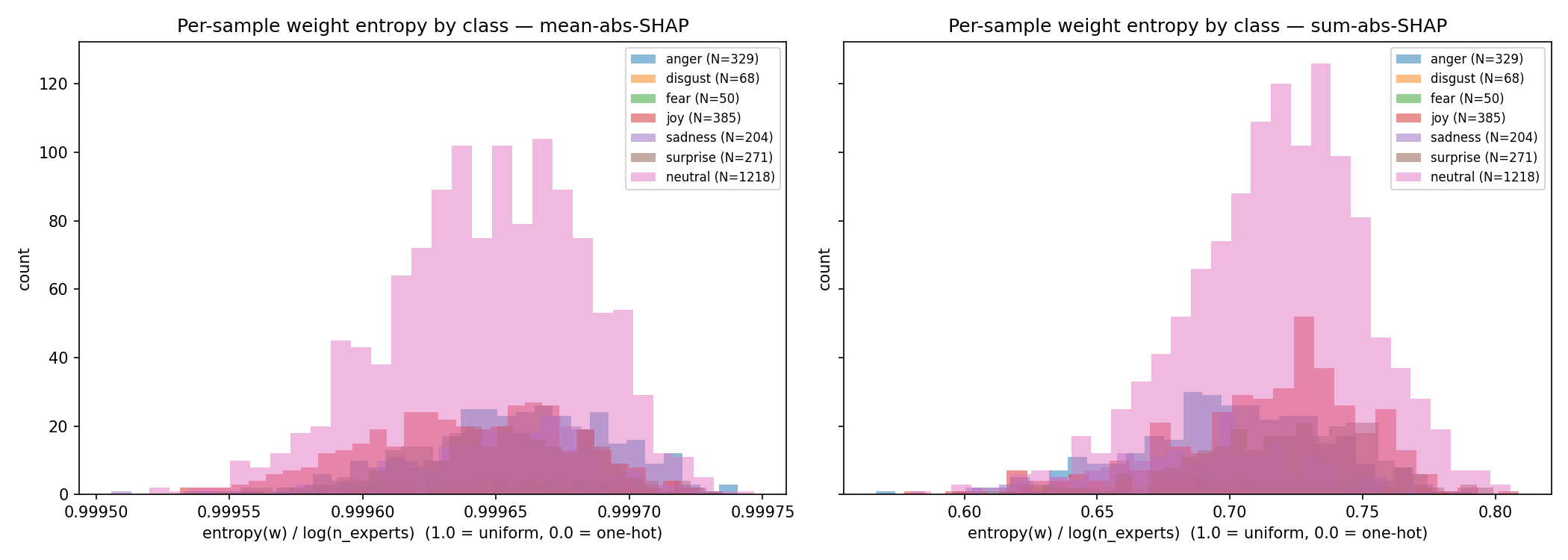}
\caption{Per-sample entropy of the expert-weight vector on MELD. Mean-abs reduction collapses to near-uniform weights. Sum-abs reduction lowers entropy, but the class distributions still overlap strongly and the trimodal expert remains dominant.}
\label{fig:entropy}
\end{figure}

The most important diagnostic result is shown in Figs.~\ref{fig:weights_mean} and~\ref{fig:weights_sum}. With the original mean-abs reduction, per-class average weights are almost uniform across the seven experts. The median-abs ablation confirms the same behaviour quantitatively: its average weight entropy is 1.9459, essentially the same as mean-abs (1.9454), and no expert dominates. With the sum-abs reduction, weights become strongly concentrated on the trimodal expert, with only a small residual weight assigned to the text-voice expert and other experts. The per-sample entropy analysis in Fig.~\ref{fig:entropy} leads to the same conclusion. Sum-abs reduces entropy to 0.3912 and assigns the argmax expert to the trimodal expert for all MELD test samples.

This diagnostic result changes the interpretation of the method. The sum-abs gate is empirically useful, but the present formulation should not be described as a strongly adaptive per-sample router. It is better described as a SHAP-weighted expert mixture in which cross-modal expert dominance is selected automatically under the evaluated conditions.

\section{Discussion}\label{sec:discussion}

\subsection{What the results support}

The results support three conclusions. First, late fusion by probability averaging is not sufficient in this setting. It underperforms early fusion by 14.2 percentage points in \wf{} on MELD with the Transformer face aggregator and by 7.89 percentage points on CMU-MOSEI. Second, adding cross-modal experts to a modular expert pool recovers early-fusion-level performance. Third, if SHAP attributions are used as a gate signal across experts of unequal dimensionality, the reduction rule must be stated and justified. Mean-abs asks how strong the average feature is, median-abs asks how strong the typical robust feature is, and sum-abs asks how much total evidence the expert carries. The new median-abs ablation shows that robustness to outlying attributions is not enough: median-abs behaves almost like mean-abs and remains about 2.74 percentage points below sum-abs in \wf{}. For the present architecture, preserving total attribution mass with sum-abs is therefore more suitable than either mean-abs or median-abs.

\subsection{What the results do not support}

The results do not support a strong claim of rich adaptive routing across emotions or samples. The diagnostic logs show that mean-abs and median-abs weighting are almost uniform and that sum-abs weighting is dominated by the trimodal expert. Therefore, the performance gain over late fusion should not be attributed primarily to a sophisticated per-sample routing mechanism. It is more accurately attributed to the inclusion of cross-modal experts combined with a reduction rule that allows those experts to receive adequate weight.

\subsection{Practical implication}

For deployment, the four-expert XGAF-Lite configuration is the most attractive variant in the current experiments. It keeps the modular unimodal predictors, includes a trimodal expert to capture cross-modal interactions, and matches early-fusion performance on MELD. The full seven-expert pool adds complexity without improving \wf{} in the reported MELD ablation. This suggests that, in settings where computational cost and interpretability matter, a small expert pool may be preferable to a larger one.

\subsection{Position relative to current MER trends}

Recent MER surveys describe a shift toward transformer-based cross-modal attention, contrastive alignment, foundation-model encoders, dynamic modality weighting, and explicit robustness mechanisms for incomplete or noisy modalities~\cite{yazici2026survey}. The present work does not compete with those end-to-end neural architectures. Its role is complementary: it asks whether a lightweight, tree-based, and explanation-derived fusion layer can recover the benefit of cross-modal modelling while preserving expert modularity. The results suggest that this is possible under clean benchmark conditions, but they also show that explanation-derived weights may collapse to cross-modal expert dominance. This observation gives a practical diagnostic criterion for future adaptive MER systems: performance should be accompanied by weight-diversity, entropy, and missing-modality analyses, not by aggregate accuracy alone.

\section{Limitations and Future Work}\label{sec:limitations}

Several limitations should be emphasised. First, the experiments use pre-extracted features rather than end-to-end fine-tuning. This makes the comparison reproducible and computationally manageable, but it limits direct comparison with the strongest neural architectures, including transformer- and foundation-model-based MER pipelines. Second, the MELD setup is per-utterance and does not model dialogue context, speaker state, or conversation graphs. Dialogue-aware models may achieve higher benchmark scores because they solve a richer temporal problem. Third, the CMU-MOSEI improvement over early fusion is statistically significant but numerically small. It should be replicated with additional seeds, bootstrap confidence intervals, and additional datasets such as IEMOCAP before being interpreted as a robust practical improvement. Fourth, the present experiments do not include missing-modality or noisy-modality stress tests, although recent MER literature treats robustness to incomplete, degraded, or asynchronous modalities as a core open problem~\cite{yazici2026survey}. Fifth, the current SHAP gate does not produce diverse per-sample routing. Sixth, the median-abs ablation has now been evaluated only for the MELD Transformer setting; it should still be repeated for the other aggregators, CMU-MOSEI, and additional seeds before broader claims about robust SHAP reductions are made. Future work should therefore test calibrated gates, hard routing, learned temperature functions, per-class priors, entropy regularisation, contrastive alignment losses, uncertainty-aware decision layers, and explicit robustness objectives. A particularly important next experiment is to evaluate whether unimodal and bimodal experts become useful when one modality is deliberately masked, corrupted, or made unavailable at test time.

\section{Conclusion}\label{sec:conclusion}

This paper revisited SHAP-weighted cross-modal expert fusion for multimodal emotion and sentiment recognition. The main empirical finding is that the choice of SHAP attribution reduction matters. Mean-abs reduction can collapse the gate toward nearly uniform weights when experts have unequal feature dimensionalities, median-abs provides a robust typical-feature alternative but empirically behaves similarly to mean-abs on MELD, and sum-abs reduction restores cross-modal expert influence and closes the performance gap to early fusion. On MELD, sum-abs XGAFv2 is statistically indistinguishable from early fusion and significantly better than naive late fusion. On CMU-MOSEI, it produces a small statistically significant improvement over early fusion and a large improvement over late fusion. At the same time, diagnostic analysis shows that the current gate is not a rich adaptive router: its useful behaviour is largely cross-modal expert dominance. The paper therefore argues for a restrained interpretation of the method. Its value lies in showing how attribution reduction and expert-pool design affect modular multimodal fusion, and in identifying the conditions under which the method behaves like early fusion rather than a genuinely adaptive routing system. In the broader MER landscape, this supports a more diagnostic style of evaluation in which accuracy is reported together with fusion behaviour, attribution scale, expert diversity, and robustness-oriented tests.

\bmhead{Acknowledgements}
The authors should insert funding, supervision, institutional, and computational-resource acknowledgements here, if applicable.

\bmhead{Use of generative AI}
An initial draft of this manuscript was prepared with assistance from a large language model. The present version was substantially revised by the authors, and the authors take responsibility for the experimental results, interpretation, references, and final scientific claims.

\bmhead{Data and code availability}
The experiments use publicly available benchmark datasets or preprocessed representations derived from them. The project code and reproducibility scripts should be released with the preprint or made available in a public repository, subject to the licenses of the underlying datasets.

\bmhead{Competing interests}
The authors declare that they have no competing interests. Replace this statement if needed.

\bibliography{references}

\end{document}